\documentclass[10pt,twocolumn,letterpaper]{article}

\usepackage{cvpr}      % To produce the REVIEW version
\usepackage[accsupp]{axessibility}
\definecolor{cvprblue}{rgb}{0.21,0.49,0.74}
\usepackage[pagebackref,breaklinks,colorlinks,allcolors=cvprblue]{hyperref}

\def\paperID{20250} % *** Enter the Paper ID here
\def\confName{CVPR}
\def\confYear{2026}

\title{
\textbf{Reasoning for Mobile User Experience with Multimodal LLMs: Task, Benchmark, and Approach}}

\author{
Ruichao Mao, Zhou Fang, Teng Guo, Hao Yang, Yaping Li, Shaohua Peng, Maji Huang, \\[-0.3ex]
Xiaoyu Lin, Shuoyang Liu, Xuepeng Li, Yuyu Zhang, Hai Rao \\[0.8ex]
\normalsize Ant Group \\[0.5ex]
\tt\small \{maoruichao.mrc, yingnan.fz, gt447200, charles.yh, pinky.lyp, shaohua.psh\}@antgroup.com\\[-0.3ex]
\tt\small \{huangmaji.hmj, axiao.lxy, liushuoyang.lsy, healy.lxp, yuyu.zyy, raohai.rh\}@antgroup.com
}

\begin{document}
\maketitle
\begin{abstract}
\textit{User experience (UX) centered on usability, perceived consistency, and functional clarity is fundamental to real-world user interfaces (UI). The application of multimodal large language models (MLLMs) in the field of user interfaces is evolving rapidly, such as visual element grounding, graphical user interface (GUI) agents, and design-to-code generation. However, research efforts on evaluating UX based on UI screenshots are still immature. To address this, we propose UXBench, a novel multimodal benchmark consisting of 2,000 VQA data samples designed to assess MLLMs' ability to perform UI-based reasoning. UXBench includes 8 tasks based on real-world UI screenshots that require fine-grained diagnosis of UX issues across layout relationships, visual hierarchy, and content consistency. Our extensive evaluation of mainstream MLLMs shows that they remain fundamentally limited in their capacity for UI-based reasoning. The results underscore the need for further advancements in this area. To bridge this gap, we propose UI-UX, an MLLM based on Qwen3-VL-4B-Thinking foundation model and enhanced via reinforcement learning with two key innovations: a reward routing mechanism that dynamically balances perceptual understanding and logical reasoning during inference, and an asymmetric transition reward that suppresses redundant or insufficient reasoning steps. Experiments demonstrate that UI-UX achieves state-of-the-art (SOTA) performance on UXBench, attaining an accuracy of 0.7963—surpassing Claude-4.5-Sonnet’s 0.6550—while exhibiting strong generalization across diverse UI tasks and maintaining low inference latency.}
\end{abstract}    
\section{Introduction}

In modern human-computer interaction systems, the determinant of success has shifted from mere functional implementation to the quality of \textbf{User Experience (UX)}\cite{ahmed2025rolelargelanguagemodels}. While UI design focuses on visual and interactive components such as layout, typography, and screen elements, UX design covers the entire user journey—emotional, cognitive, and behavioral responses before, during, and after interaction\cite{bharath2023leader}.  

The rise of large language models (LLMs) and multimodal large language models (MLLMs) has greatly advanced UX in human-computer interaction, enabling UI understanding~\cite{you2024ferret,lu2025guiodyssey,lin2025showui}, UI generation~\cite{yang2025uiugunifiedmllmui,lu2023uilayoutgenerationllms,duan2023towards,ran2024guardian}, and user affect recognition~\cite{zhao2025llms,sun2025dialoguemllm,lian2025affectgpt} from raw visual or multimodal inputs. These models provide capabilities such as pixel-level interface parsing, intent-driven UI synthesis, and emotion-aware interaction modeling, positioning foundation models as scalable infrastructure for data-driven, user-centric interface systems.

However, UX issues often arise from misalignments between design conventions and user mental models, rather than visible layout defects. For example, a modal dialog occluding navigation bars or inconsistencies between advertised and actual content may appear visually correct yet result in frustration, eroded trust, and user churn. This shift from ``perceiving interfaces'' to ``inferring experiences'' poses fundamental challenges for multimodal models aiming at automated UX evaluation.

Existing UI-based benchmarks such as Screen2Words \cite{wang2021screen2wordsautomaticmobileui}, Mobile-bench \cite{deng2024mobile}, and VisualWebBench \cite{liu2024visualwebbenchfarmultimodalllms} predominantly evaluate visual perception tasks (e.g., caption generation, element detection, layout parsing) rather than \textbf{UX reasoning}. These benchmarks lack objectives grounded in behavioral outcomes, cognitive psychology, and trust dynamics, making them insufficient for detecting design patterns that trigger user errors or negative emotions. Consequently, current MLLMs remain limited in applications such as UI design assessment, automated testing, and intelligent design assistance.

\begin{figure*}[t]
\centering
\includegraphics[width=1.02\textwidth, clip, trim=100 546 100 125]{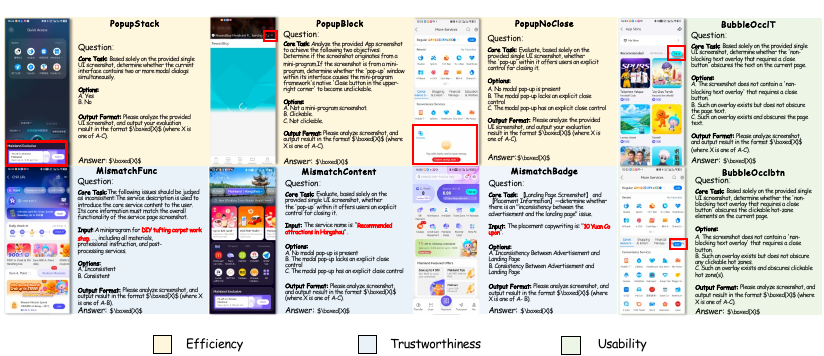} 
\vspace{-0.3em} 
\caption{\textbf{UXBench samples spanning Efficiency, Trustworthiness, and Usability dimensions.} Each case requires visual-semantic reasoning to detect UX issues (e.g., overlapping modals, deceptive content, missing controls). Red bounding boxes indicate ground-truth defect regions for quantitative assessment. Tasks involve inferring experiential consequences beyond pixel-level perception.} 
\label{fig:UXBenchSamples}
\vspace{-0.8em} 
\end{figure*}

Recent work has explored some UI-based tasks such as design-to-code generation\cite{yang2025uiugunifiedmllmui}, automated GUI operation, and component recognition. Yet these approaches typically rely on manually labeled instruction--action pairs or oversimplify UX evaluation into binary ``good/bad'' classification, ignoring the multi-dimensional, context-dependent nature of UX issues. On complex cases(e.g., nested pop-ups making operations irreversible, mismatched service names and functionalities) current models often produce vague, erroneous, or contradictory reasoning. This reveals deficiencies in semantic association, counterfactual reasoning, and intent modeling; aligning vision and text alone is insufficient for genuine UX understanding.

To address these limitations, we introduce \textbf{UXBench}, the first multimodal benchmark for evaluating MLLMs' UI-UX reasoning capabilities. UXBench contains over 2,000 real UI screenshots with user feedback, categorizing UX issues into three dimensions: usability, efficiency, and trustworthiness, and further into 8 fine-grained diagnostic tasks (See Fig. \ref{fig:UXBenchSamples}). Each task is posed as a two- or three-choice question requiring causal reasoning and mapping to design principles rather than keyword matching. Large-scale MLLM-assisted annotation, combined with two rounds of expert validation by four senior UX specialists, ensures high label consistency.

We further propose \textbf{UI-UX}, a reinforcement learning-based MLLM enhancement framework built upon Qwen3-VL-4B-Thinking\cite{Qwen-VL,Qwen2-VL,Qwen2.5-VL}. UI-UX adopts task-adaptive reward routing with accuracy rewards for UX diagnostic tasks, ROUGE-L scores for semantic alignment in general UI understanding, and hit rewards for grounding tasks. Additionally, we introduce an Asymmetric Transition Reward mechanism to suppress redundant reasoning and reduce inference latency. Without manual preference annotation, the model is optimized end-to-end via the GRPO\cite{shao2024deepseekmathpushinglimitsmathematical} algorithm, achieving 73.8\% accuracy on UXBench and demonstrating strong cross-domain generalization.

\noindent
\textbf{Our contributions are as follows:}
\begin{itemize}
    \item \textbf{First UX reasoning benchmark}: We propose UXBench, defining 8 fine-grained, evaluable, user-centered UI diagnostic tasks, filling a critical gap in the field.
    \item \textbf{Reward routing}: We introduce hard negative sampling and semantic preservation augmentation strategies to address positive sample scarcity and extreme class imbalance in real scenarios.
    \item \textbf{Efficient reasoning}: The proposed UI-UX model combines reward routing and overthinking penalty, achieving MLLM UX reasoning performance surpassing human experts, with low latency and high robustness.
\end{itemize}

\section{Related Work}

\subsection{UI-based Benchmarks}
Recent benchmarks have focused on evaluating models' visual understanding of user interfaces (UI). Screen2Words  \cite{wang2021screen2wordsautomaticmobileui} and GUI-Text \cite{cui2024guitext} train models to generate UI descriptions from paired screenshot-text data, but are limited to static element recognition and naming, without modeling interaction logic or user intent.  RICO \cite{rico} and VisualWebBench \cite{liu2024visualwebbenchfarmultimodalllms} further introduce layout parsing and element localization tasks, remaining within the ``perception'' layer—detecting buttons, reading text, and extracting structure—rather than judging whether a design may confuse users. None of these benchmarks defines evaluation dimensions related to User Experience, and thus cannot measure reasoning such as whether a pop-up obscures critical operations or a badge misleads users. Our work fills this gap by constructing the first benchmark centered on fine-grained UX diagnostics, enabling models to move from understanding interfaces to understanding users.

\subsection{UX Issue and Detection Methods}
User Interfaces (UIs) often suffer from issues such as text overlap, missing images, and layout distortion arising from device diversity and design complexity, undermining usability and accessibility. Vision-based tools like OwlEyes-online~\cite{su2021owleyes} and Nighthawk~\cite{liu2022nighthawk} can localize visual bugs, while Metamorphosis~\cite{su2022metamorphosis} detects scaling defects via metamorphic testing. However, most taxonomy-based approaches address only predefined issue types in limited or synthetic datasets, overlooking higher-level design smells and cross-component inconsistencies. UISGPT~\cite{yang2024uisgpt} applies large language models to identify guideline violations with explanations, but struggles with complex, context-sensitive reasoning. Our work differs by constructing a rich, real-world dataset that captures both visual and structural UX issues, and by employing MLLMs with causal reasoning to detect and explain complex design flaws.

\subsection{Reasoning in MLLMs}
% Reasoning-capable MLLMs have advanced complex visual reasoning by introducing explicit \emph{Chain-of-Thought} (CoT) generation and self-consistency verification. Examples include LLaVA-1.5 ~\cite{liu2024improvedbaselinesvisualinstruction} for multi-step image reasoning, MiniGPT-4-v2 \cite{zhu2023minigpt4enhancingvisionlanguageunderstanding} and InternVL \cite{chen2024internvl,zhu2025internvl3, wang2025internvl3_5} for instruction-tuned logical scene analysis, and CogVLM \cite{} and Visual-CoT \cite{visualcot} for structured output reasoning in tasks like MMMU and ChartQA. Despite these advances, existing models focus on fact-based reasoning (e.g., object relations, physics consistency) rather than \emph{experience-based reasoning}, and cannot evaluate if a button placement violates user habits, if a pop-up causes anxiety, or if a description misleads trust. Such UX reasoning demands counterfactual inference, design principle grounding, and mental model alignment. Our work is the first to embed reasoning into the human–computer interaction context, enabling MLLMs to perform user-centric causal reasoning in UI scenarios.

Reasoning-capable MLLMs have advanced complex visual reasoning by introducing explicit \emph{Chain-of-Thought} (CoT) generation and self-consistency verification. Examples include LLaVA-1.5~\cite{liu2024improvedbaselinesvisualinstruction} for multi-step image reasoning, MiniGPT-4-v2~\cite{zhu2023minigpt4enhancingvisionlanguageunderstanding} and InternVL~\cite{chen2024internvl,zhu2025internvl3,wang2025internvl3_5} for instruction-tuned logical scene analysis, and CogVLM and Visual-CoT~\cite{shao2024visualcot} for structured output reasoning in tasks like MMMU and ChartQA. Recent works such as GRPO-$\lambda$~\cite{parthasarathi2025grpolambdacreditassignmentimproves}, Step Pruner~\cite{wu2025tokenlengthsteppruner}, and CoRE-Eval~\cite{Zhao_2025} address reasoning efficiency by pruning redundant steps and assessing step-level importance, providing insights for reducing latency while preserving accuracy. Despite these advances, most models still emphasize fact-based reasoning over \emph{experience-based reasoning}, leaving them unable to identify UX issues such as inappropriate button placement, disruptive pop-ups, or misleading descriptions. Our work is the first to embed such reasoning into the human–computer interaction context, enabling MLLMs to perform user-centric causal inference in UI scenarios.

\section{UXBench}

Existing vision–language benchmarks mainly address general scene understanding and image generation, with limited evaluation of models' reasoning and comprehension capabilities in real-world UI. To address this gap, we present \textbf{UXBench}—the first vision–language benchmark for UX defect diagnosis—designed to systematically assess multimodal large language models (MLLMs) on fine-grained, higher-order reasoning within realistic UI scenarios.

\subsection{Task Definition}
Traditional UI recognition focuses on visual perception, detecting and classifying visible interface components from screenshots and reconstructing their layout. These methods rely on explicit feature extraction, producing outputs derivable directly from pixel information—without modeling user behavior or intent.  
In contrast, UX diagnosis requires identifying components and evaluating whether their arrangement, interaction logic, and semantics pose potential experience risks. Such issues often stem from mismatches between design conventions and human usage habits. For example, assessing whether “a popup occludes a button” involves detection, spatial analysis, and causal inference about operational impact—reflecting reasoning beyond pixel-level perception.

For fine-grained evaluation, we organize UX diagnosis into three dimensions operationalized from rigorous HCI frameworks:

\begin{enumerate}
 \item \textbf{Usability}: Clarity of operation and feedback.
 This dimension aligns with "Operability" \cite{weichbroth2025factorsinfluencingperceivedusability} in HCI frameworks, focusing on the visibility and accessibility of screen objects.
 \begin{enumerate}[label=(\alph*)]
 \item BubbleOcclT: Textual overlay occluding page text.
 \item BubbleOcclBtn: Textual overlay blocking clickable elements.
 \end{enumerate}

 \item \textbf{Efficiency}: Minimizing operational and cognitive cost.
 Empirical data demonstrates that efficiency is the highest-rated usability factor (mean: 4.07/5)\cite{weichbroth2025factorsinfluencingperceivedusability}.
 \begin{enumerate}[label=(\alph*)]
 \item PopupNoClose: Popup without explicit close control.
 \item PopupBlockClose: Popup affecting native close button clickability.
 \item PopupStack: Multiple modal popups present simultaneously.
 \end{enumerate}

 \item \textbf{Trustworthiness}: Maintaining consistency and credibility.
 This dimension operationalizes "Persuasiveness" and "Security"\cite{brangier2018usabilityuserexperience} from HCI frameworks.
 \begin{enumerate}[label=(\alph*)]
 \item MismatchBadge: Badge content inconsistent with landing page.
 \item MismatchContent: Service name inconsistent with page text.
 \item MismatchFunc: Description inconsistent with provided functionality.
 \end{enumerate}
\end{enumerate}

\subsection{Data Pipeline}
UXBench is built through a multi-stage pipeline combining large-scale real user feedback, MLLM-assisted annotation, and expert quality control.  

\textbf{Raw data collection}: screenshots and textual descriptions from in-app feedback across diverse mobile application scenarios.  

\textbf{Relevance filtering}: Gemini-2.5-Pro classifies relevant feedback; a fine-tuned Qwen3-VL-2B replicates decisions at scale to retain high-confidence UX samples.  

\textbf{Core categorization}: relevant samples are labeled by Gemini into dimensions and subtasks using few-shot prompting, with multi-round voting for ambiguous cases.  

\textbf{Human verification}: Four senior UX researchers conduct two-stage manual validation, beginning with independent annotation by all four researchers followed by cross-validation with disagreement resolution.

\textbf{Final dataset}: balanced sampling from positive and negative pools yields 2,000 high-quality image–question pairs covering typical issues and normal UI scenarios.

% \begin{figure}[t]
% \centering
% \includegraphics[width=1.0\columnwidth, trim=350 200 350 230, clip]{img/fenbu.pdf} 
% \caption{Data distribution across subtasks in UXBench.}
% \vspace{-0.8em} 
% \label{fig:fenbu}
% \end{figure}

\begin{figure}[t]
\centering
\includegraphics[width=1.0\columnwidth, trim=200 520 170 200, clip]{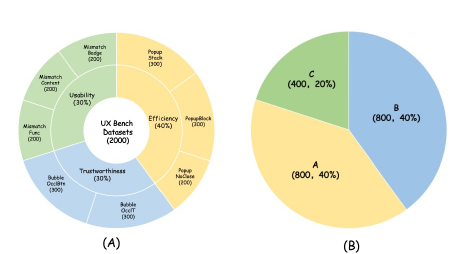} 
\caption{{\textbf{Data distribution in UXBench.} 
    (A) Distribution across different subtasks in the benchmark dataset (2000 samples total). 
    (B) Distribution of user interaction options}}
\vspace{-0.3em} 
\label{fig:benchdata}
\end{figure}

\subsection{Data Distribution}
UXBench emphasizes diversity across tasks and option distribution. (See Fig \ref{fig:benchdata})

\textbf{Tasks}: three dimensions with 2–3 subtasks each, averaging 200–300 instances; positive and negative cases are balanced.  

\textbf{Options}: 40\% with 2 choices, 60\% with 3 choices.  

\textbf{Coverage}: includes iOS/Android platforms, light/dark themes, and multiple orientations for better generalization.

\section{UI-UX}
\label{sec:formatting}
To bridge the significant gap between existing Multimodal Large Language Models (MLLMs) and human experts in user interface (UI) reasoning, we introduce UI-UX, a reinforcement learning-enhanced model specifically optimized for UX diagnostic tasks. Built upon the Qwen-VL-4B-Thinking architecture, UI-UX employs task-aware reinforcement learning (RL) to provide end-to-end guidance for the reasoning generation process, achieving substantial improvements in reasoning efficiency while maintaining diagnostic accuracy. 

%------------------------------------------------------------------------

\subsection{Training Data Composition}
\begin{figure}[t]
\centering
\includegraphics[width=1.0\columnwidth, clip, trim=210 535 220 208]{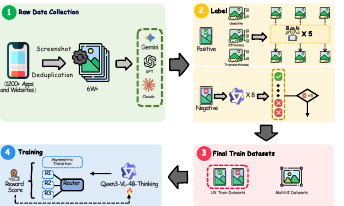}
\caption{\textbf{UI-UX training pipeline overview.} \textbf{(1) Raw Data Collection:} 6M+ screenshots from 1,200+ apps/websites, deduplicated via pHash. \textbf{(2) Label Generation:} MLLM-based pseudo-labeling with 5× positive augmentation and 8× hard negative mining. \textbf{(3) Final Datasets:} 21,761 UX samples across eight tasks + 4,919 MultiUI samples for regularization. \textbf{(4) Training:} RL optimization with asymmetric transition reward and reward router on Qwen3-VL-4B-Thinking.}
\label{fig:pipeline}
\end{figure}

\textbf{Dataset Collection and Annotation.}  
We collect a large-scale dataset of real-world UI screenshots via automated testing scripts executed on eight physical devices spanning Android, HarmonyOS, and iOS. By systematically navigating core user flows in 1,200+ popular applications and webpages, we gather 832,432 raw screenshots(See Fig \ref{fig:pipeline}). To mitigate visual redundancy caused by deterministic execution, we apply perceptual hashing (pHash) and retain only one sample per group of images with Hamming distance $\leq 5$, where the Hamming distance between two pHash ($n$~bit) vectors $\mathbf{h}_i, \mathbf{h}_j \in \{0,1\}^n$ is defined as:

\begin{equation}
d_H(\mathbf{h}_i, \mathbf{h}_j) = \sum_{k=1}^{n} \mathbb{I}\big[h_i[k] \neq h_j[k]\big]
\label{eq:hamming}
\end{equation}

and $\mathbb{I}[\cdot]$ denotes the indicator function. This yields 68,138 unique UI screenshots.  
For scalable, high-fidelity labeling, we propose a two-stage annotation pipeline. First, we leverage multiple MLLMs (GPT5, Gemini2.5-pro, Claude4.5) to predict UX issues across three dimensions (Usability, Efficiency, Content Consistency) and eight sub-tasks, using carefully engineered prompts. Predictions are aggregated via majority voting to generate robust pseudo-labels. Second, all pseudo-labels are manually verified and corrected by five experienced UX researchers, ensuring label accuracy.

\textbf{Positive--Negative Sample Balancing.}  
UX issue detection suffers from severe class imbalance where positive samples (defective UIs) are heavily outnumbered. This causes models to bias toward the majority class and makes GRPO training degenerate when all samples in a group share the same label.

We apply hard negative mining by sampling each negative image 8 times with Qwen3-VL-Thinking-4B (temperature = 1.0). Only samples with inconsistent predictions ($\leq 5/8$ votes) are retained as hard negatives, filtering out easy cases and ensuring challenging training examples.

\textbf{Mixed-Task Regularization.}  
Training exclusively on the UX issue dataset triggers catastrophic forgetting, eroding the model's general UI understanding and degrading performance. To mitigate this, we incorporate 4,919 randomly sampled examples from the MultiUI dataset \cite{liu2024harnessing}---a large-scale corpus of 7.3M UI samples.

\textbf{Final Training Data Composition.}
Our final training dataset contains 26,680 samples strategically distributed across UX-specific and multi-domain data. 

%-------------------------------------------------------------------------
\subsection{Reward Design}
Our reward function is formulated as:
\begin{equation}
\mathcal{R} = \mathcal{R}_{\text{routing}} + \mathcal{R}_{\text{transition}}
\label{eq:reward_total}
\end{equation}

where $\mathcal{R}_{\text{routing}}$ is a task-adaptive reward that dynamically 
selects an accuracy-oriented metric according to the downstream task 
(e.g., UX reasoning accuracy, OCR edit distance, or grounding IoU), 
and $\mathcal{R}_{\text{transition}}$ encourages concise reasoning by penalizing 
redundant or superfluous inference steps. Empirical results demonstrate 
that incorporating $\mathcal{R}_{\text{transition}}$ significantly improves both 
training efficiency and inference speed(FLOPs and latency), 
without compromising task performance, thereby promoting succinct 
yet accurate multimodal reasoning.

% Reward Routing ROUGE-L
\textbf{Reward Routing.} We propose a task-aware reward routing mechanism that dynamically selects the most appropriate reward function based on the origin and format of each training sample, effectively decoupling optimization objectives across heterogeneous multimodal tasks. For UX issue detect samples, typically presented as multiple-choice questions requiring precise mathematical or logical reasoning. we adopt answer accuracy (MathAcc) as the reward, computed via robust LaTeX parsing and semantic equivalence verification to determine whether the model’s final prediction matches the ground-truth solution. For MultiUI general understanding samples, where the model is required to generate natural-language descriptions of UI elements or scenes, we employ ROUGE-L to quantify textual fidelity, defined as:
\begin{equation}
\mathcal{R}_{\text{ROUGE-L}} = \frac{(1 + \beta^2) \cdot P_{\text{LCS}} \cdot R_{\text{LCS}}}{\beta^2 \cdot P_{\text{LCS}} + R_{\text{LCS}}}
\label{eq:rouge}
\end{equation}

where $P_{\text{LCS}}$ and $R_{\text{LCS}}$ denote precision and recall based on LCS.

% Visual grounding reward
For visual grounding tasks:
\begin{equation}
\mathcal{R}_{\text{hit}} = \mathbb{I}\left( (x_c^{\text{pred}}, y_c^{\text{pred}}) \in [x_1^{\text{gt}}, x_2^{\text{gt}}] \times [y_1^{\text{gt}}, y_2^{\text{gt}}] \right)
\label{eq:hit}
\end{equation}

%-------------------------------------------------------------------------
\textbf{Asymmetric Transition Reward.} Transition markers indicate critical reasoning transitions within chains. Motivated by the correlation between transition markers and reasoning quality, we design an asymmetric reward function to balance reasoning sufficiency and conciseness.
Given a model prediction $p$ with transition marker count $T(p)$ and ground truth label $y$, 
we define the correctness indicator $\mathbf{1}_{p = y}$.
The complete reward function is formulated as:
The reward function is:
\begin{equation}
\begin{aligned}
R(p, y) &= 1_{p=y} \cdot R_{\text{correct}}\big(T(p)\big) \\
&\quad+ \big(1 - 1_{p=y}\big) \cdot R_{\text{incorrect}}\big(T(p)\big)
\end{aligned}
\label{eq:asym_reward}
\end{equation}

Where the two branch functions are defined as:
\begin{equation}
\begin{aligned}
R_{\text{correct}}(T) &= \max\big(r_{\text{base}} - \alpha \cdot T, \; r_{\min}\big) \\
R_{\text{incorrect}}(T) &= \min\big(\alpha \cdot T, \; r_{\max}\big)
\end{aligned}
\label{eq:branch_rewards}
\end{equation}
Here $r_{\text{base}} = 1.0$ denotes the base reward, $\alpha$ is the penalty coefficient, and $r_{\min}$, $r_{\max}$ represent the reward boundaries for correct and incorrect predictions, respectively. This design embodies three core mechanisms: (1) Penalty for correct predictions: reward decreases linearly with transition markers, discouraging over-deliberation after reaching correct answers, while the lower bound $r_{\min}$ ensures that even verbose correct answers outperform incorrect ones; (2) Exploration incentive for incorrect predictions: reward increases linearly with transition markers, encouraging more thorough reasoning exploration when uncertain, while the upper bound $r_{\max}$ prevents exploiting the reward by simply adding redundancy; (3) Correctness-first guarantee: the constraint $r_{\min} > r_{\max}$ mathematically ensures that any correct answer strictly yields higher reward than any incorrect answer.

The key innovation of this design lies in establishing an insurmountable reward gap $\mathcal{G}$ that mathematically enforces the correctness-first principle:
\begin{equation}
\mathcal{G} = \inf_{T \ge 0} R_{\text{correct}}(T) - \sup_{T \ge 0} R_{\text{incorrect}}(T) = r_{\min} - r_{\max} > 0
\label{eq:reward_gap}
\end{equation}
This strictly positive reward gap guarantees that for any transition marker counts $T_1, T_2 \geq 0$, we have $R_{\text{correct}}(T_1) > R_{\text{incorrect}}(T_2)$. During policy optimization, let $\pi_\theta$ denote the parameterized policy. The expected reward naturally decomposes as:
\begin{equation}
\begin{aligned}
\max_{\theta} \ \mathbb{E}_{\pi_\theta}[R] &= \max_{\theta} \Big( \mathbb{E}_{\pi_\theta}[1_{p=y}] \cdot \mathbb{E}[R_{\text{correct}} \mid p = y] \\
&\quad+ \big( 1 - \mathbb{E}_{\pi_\theta}[1_{p=y}] \big) \cdot \mathbb{E}[R_{\text{incorrect}} \mid p \neq y] \Big)
\end{aligned}
\label{eq:policy_opt}
\end{equation}

Since $\mathcal{G} > 0$, the gain from improving accuracy 
$\mathbb{E}_{\pi_\theta}\big[ \mathbf{1}_{p = y} \big]$ 
always dominates conciseness optimization, theoretically preventing the degenerate solution of "sacrificing correctness for conciseness." 

For transition marker counting, we employ structured matching: let $\mathcal{V}_t$ denote the transition marker vocabulary and $\mathcal{P}$ the position prefix set (e.g., "\textbackslash n", "--"). For text $p = \{ w_1, \dots, w_n \}$, we define
\begin{equation}
T(p) = \sum_{i=2}^n \mathbf{1}_{w_i \in \mathcal{V}_t} \cdot \mathbf{1}_{\exists k \in \mathcal{P} : w_{i-1} = k}
\end{equation}
The position constraint filters out casual in–sentence transitions, capturing only explicit logical transitions in the reasoning path.
%-------------------------------------------------------------------------
\subsection{Training Details}
We train all models using GRPO, a sample-efficient on-policy RL algorithm that stabilizes policy updates via per-token reward weighting and group-wise normalization. Training is conducted over 130 hours on a 16-PPU cluster using Qwen3-VL-4B-Thinking as the base model. We apply LoRA with rank $r = 8$ and scaling $\alpha = 32$ to all linear layers, while freezing the vision encoder (ViT) and vision--language aligner to preserve pretrained representations. Training uses DeepSpeed ZeRO-2 for memory efficiency, and vLLM with tensor parallelism (4 GPUs per node) and 50\% GPU memory utilization to enable high-throughput rollout generation. Maximum sequence length is 16K tokens, with a completion length cap of 8K to support extended reasoning chains. Each batch contains 2~samples per GPU (effective batch size $= 32$), and 16~responses are generated per prompt during rollouts using $\mathrm{temperature=1.0}$, $\mathrm{top-}$$k = 80$, and $\mathrm{top-}$$k = 1.0$. We optimize with a learning rate of $2\times 10^{-6}$, linear warmup over 5\% of steps, followed by cosine decay over one epoch.
\begin{figure}[t]
\centering
\includegraphics[width=1.0\columnwidth, clip,trim=100 20 95 100]{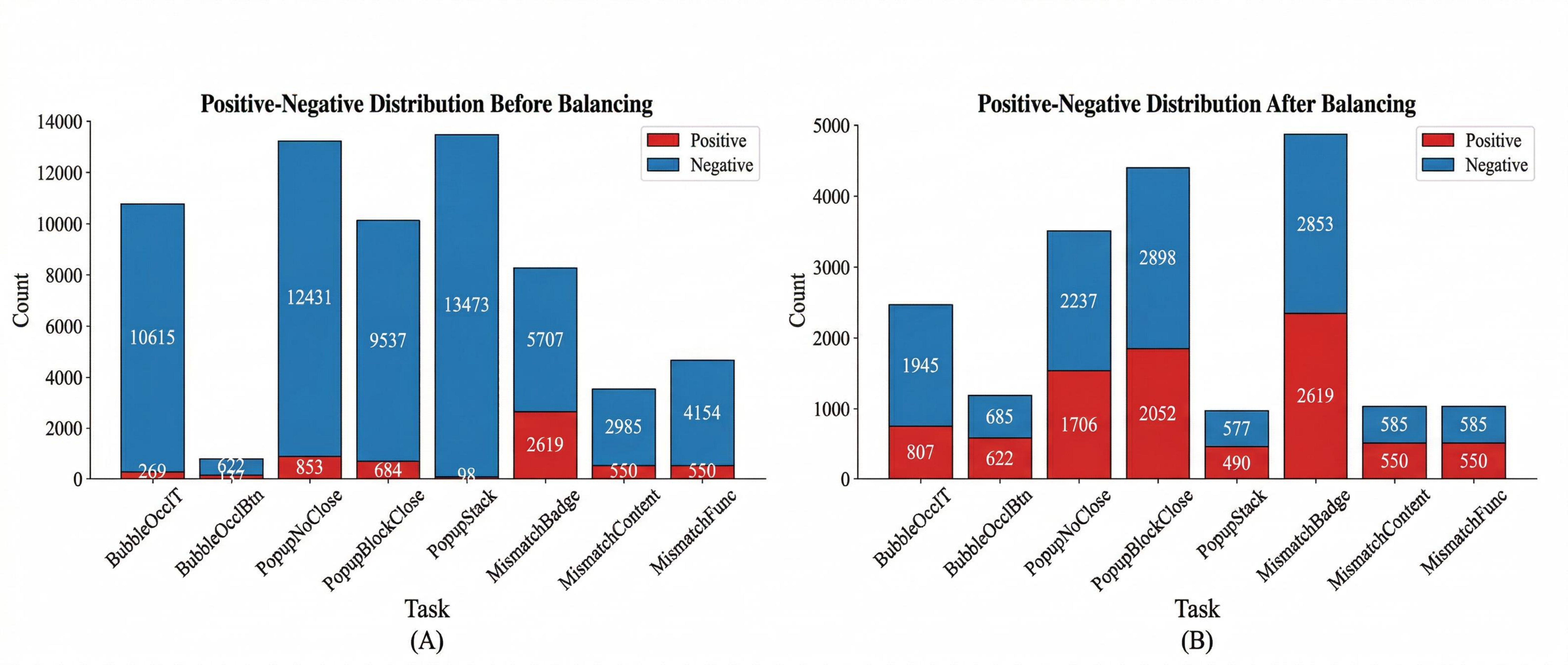} 
\caption{\textbf{Positive-negative distribution before and after balanced sampling.} (a)  Original data shows severe class imbalance with positive samples (red) heavily outnumbered by negative samples (blue). (b) After applying hard negative mining and positive augmentation, the dataset achieves improved balance across all eight tasks.}
\label{fig:Before_sample_distribution}
\end{figure}

\begin{table*}[htbp]
\renewcommand{\arraystretch}{0.9} % 行高
\footnotesize
\centering
\caption{Performance comparison of different models on UXBench. Reasoning models generally outperform instruct models across tasks. $^*$ indicates format parsing failures or excessively long reasoning outputs during evaluation. Bold values represent the best performance across all models for each metric.}
\label{tab:model_performance_reduced}
\begin{tabular}{
    p{3cm}  % Model
    p{0.9cm}  % Params
    p{0.9cm} p{0.9cm}  % Usability
    p{0.9cm} p{0.9cm} p{0.9cm}  % Efficiency
    p{0.9cm} p{0.9cm} p{0.9cm}  % Trustworthiness
    p{1cm}  % AVG
}
\toprule
Model & Params &
Bubble OccT & Bubble OccBtn &
Popup No Close & Popup Block Close & Popup Stack &
Mismatch Badge & Mismatch Content & Mismatch Func &
AVG. \\
\midrule
\multicolumn{11}{l}{\textbf{Instruct Model}} \\
\midrule
Llava3-Next \cite{liu2024llavanext}  & 8B& 0$^*$ & 0.1 & 0.22 & 0.22 & 0.26 & 0.19 & 0$^*$ & 0$^*$ & 0.1200 \\
Qwen2.5-VL \cite{Qwen2.5-VL}& 72B & 0.09$^*$ & 0.51 & 0.70 & 0.55 & 0.56 & 0.69 & 0.68 & 0.71 & 0.5482 \\
Qwen3-VL \cite{Qwen3-VL}& 235B & 0.30 & 0.57 & 0.79 & 0.40 & 0.52 & 0.62 & 0.64 & 0.70 & 0.5600 \\
InternVL-3 \cite{wang2025internvl3_5} & 2B & 0.3 & 0.34 & 0.61 & 0.0 & 0.34 & 0.52 & 0.11 & 0.08 & 0.2875\\
InternVL-3.5 \cite{wang2025internvl3_5} & 2B & 0.42 & 0.34 & 0.43 & 0.26 & 0.37 & 0.61 & 0.72 & 0.76 & 0.4888\\
MiniCPM-V-4.5 \cite{yao2024minicpm} & 8B & 0.3 & 0.39 & 0.56 & 0.33 & 0.55 & 0.53 & 0.65 & 0.76 & 0.5086 \\
\midrule
\multicolumn{11}{l}{\textbf{Reasoning Model}} \\
\midrule
GLM-4.1-Thinking \cite{vteam2025glm45vglm41vthinkingversatilemultimodal} & 9B & 0$^*$ & 0.6 & 0.62 & 0.34 & 0.32 &0.32 &  0.04$^*$ & 0.01$^*$ & 0.2813 \\
Qwen3-VL-Thinking \cite{Qwen3-VL}& 4B & 0.47 & 0.49 & 0.72 & 0.39 & 0.50 & 0.52 & 0.63 & 0.65 & 0.5254 \\
Qwen3-VL-Thinking \cite{Qwen3-VL} & 235B & 0.52 & 0.61 & \textbf{0.82} & 0.47 & 0.56 & 0.54 & 0.64 & 0.70 & 0.5854 \\
InternVL-3.5 \cite{wang2025internvl3_5} & 4B & 0$^*$ & 0.39 & 0.65 & 0.38 & 0.50 & 0.59 & 0.68 & 0.72 & 0.4888\\
MimoVL-0528 \cite{mimovltechnicalreport} & 7B & 0.58 & 0.56 & 0.74 & 0.41 & 0.52 & 0.54 & 0.40 & 0.6 & 0.5438 \\
Claude-3.7-Sonnet \cite{Claude37} & -- & 0.64 & 0.66 & 0.78 & 0.56 & 0.53 & 0.61 & 0.65 & \textbf{0.76} & 0.6488\\
Claude-4-Sonnet \cite{Claude4} & -- & 0.64 & 0.52 & 0.77 & 0.46 & 0.52 & 0.68 & 0.66 & 0.74 & 0.6238\\
Claude-4.5-Sonnet \cite{Claude45} & -- & 0.65 & 0.55 & 0.77 & 0.53 & 0.66 & 0.7 & 0.66 & 0.72 & 0.6550\\
% GPT-5-chat\cite{gpt5} & -- & -- & -- & -- & -- & -- & -- & -- & -- & --\\
% Gemini-2.5-pro\cite{gemini25} & -- & -- & -- & -- & -- & -- & -- & -- & -- & --\\
\textbf{UI-UX(ours)} & 4B & \textbf{0.79} & \textbf{0.88} & 0.79 & \textbf{0.77} & \textbf{0.94} & \textbf{0.72} & \textbf{0.71} & \textbf{0.76} & \textbf{0.7963} \\
\bottomrule
\end{tabular}
% \begin{flushleft}
% \end{flushleft}
\end{table*}

\section{Experiment}
\subsection{Main Results on UXBench}

\noindent\textbf{Evaluation Setup.}
We evaluate all models under consistent conditions with temperature set to 0 and maximum generation length of 8,192 tokens to ensure fair comparison across different architectures.

\noindent\textbf{Overall Performance.}
Table~\ref{tab:model_performance_reduced} presents the comprehensive evaluation results across eight UXBench tasks spanning usability, efficiency, and trustworthiness dimensions. Our UI-UX model, with only 4B parameters, achieves the best overall performance with an average score of 0.7963, substantially outperforming both instruct and reasoning models. Specifically, UI-UX (ours) surpasses the best reasoning model Claude-4.5-Sonnet (0.6550) by 21.6\% and the top instruct model Qwen3-VL 235B (0.56) by 42.2\%, demonstrating the effectiveness of our domain-specific design on UXBench despite being significantly smaller in scale. Additionally, reasoning models generally demonstrate superior performance compared to instruct models on UXBench. For instance, Qwen3-VL-Thinking 235B (0.5854) outperforms its instruct counterpart Qwen3-VL 235B (0.56) by 4.5\%, highlighting the importance of chain-of-thought reasoning capabilities for these challenging benchmarks even when controlling for model scale.

\noindent\textbf{Overthinking Problem.}
Despite their superior accuracy, reasoning models face a significant challenge on UXBench: overthinking that causes generation to exceed the 8,192 token limit, resulting in parsing failures (marked with $^*$). This issue disproportionately affects smaller models. For instance, GLM-4.1-Thinking (9B) suffers from multiple parsing failures (Bubble OccT: 0$^*$, Mismatch Badge: 0.04$^*$, Mismatch Content: 0.01$^*$), achieving only 0.2813 average, while InternVL-3.5 (4B) fails on Bubble OccT (0$^*$) with 0.4888 overall. In contrast, larger reasoning models like Qwen3-VL-thinking (235B) and Claude-4.5-Sonnet successfully avoid these failures. This suggests that model scale is critical for controlling reasoning verbosity while preserving reasoning quality on UXBench's complex visual reasoning tasks.

\subsection{Ablation Study on Training Strategies}
To investigate the impact of different data sampling and augmentation strategies on model performance, we conduct comprehensive ablation studies on UXBench, as shown in Table~\ref{tab:ablation}. All experiments are based on Qwen3-VL-4B-Thinking.

\noindent\textbf{Baseline Performance.}
The baseline model achieves an accuracy of 52.54\% on UXBench. However, directly training with the full dataset before balancing suffers from severe reward hacking—the model exploits spurious correlations in training data, leading to inflated training metrics that fail to generalize to the evaluation set. This makes the unbalanced training configuration impractical for real-world deployment.

\noindent\textbf{Hard Negative Mining (HNM).}
Incorporating hard negative mining effectively addresses the reward hacking issue by prioritizing challenging samples that expose model weaknesses. By focusing on frequently misclassified cases (e.g., subtle modal overlaps, deceptive UI patterns), HNM guides the model to learn more discriminative decision boundaries. This strategy alone yields a substantial gain of +19.41\% (71.95\%), demonstrating its critical role in mitigating dataset bias and establishing a robust training foundation.

\noindent\textbf{Positive Sample Strategies.}
Building upon HNM, we compare two approaches for enriching positive samples. Positive resampling increases the sampling frequency of existing defect instances, achieving 75.79\% (+23.25\%). In contrast, positive augmentation generates diverse variations through layout transformations and content perturbations, reaching 77.71\% (+25.17\%). The augmentation approach outperforms resampling by +1.92\%, as it introduces greater sample diversity, exposing the model to varied visual contexts while preserving semantic defect patterns, thereby reducing overfitting to specific UI layouts.

\noindent\textbf{MultiUI Integration.}
Finally, combining HNM and positive augmentation with MultiUI training achieves the best performance at 79.63\% (+27.09\%). The MultiUI approach exposes the model to diverse interface designs, interaction paradigms, and visual styles beyond the target domain. This cross-domain training enhances generalization capability, enabling robust UX assessment across heterogeneous applications. Notably, MultiUI contributes an additional +1.92\% improvement over positive augmentation alone, validating its effectiveness in building cross-domain robustness.

The ablation study reveals three critical insights: (1) Unbalanced training causes reward hacking: under extreme imbalance, the model predicts all negatives, yielding 40\% accuracy but 0\% meaningful accuracy, while hard negative mining provides a +19.41\% gain by addressing this issue. (2) Augmentation-based diversity (+25.17\%) is more effective than resampling (+23.25\%) for positive samples, with a +1.92\% advantage. (3) Multi-domain exposure through MultiUI is essential for achieving state-of-the-art performance (79.63\%), demonstrating that generalizable UX reasoning requires both within-domain hard case coverage and cross-domain knowledge transfer.
\begin{table}[t]
\centering
\caption{Ablation study on training strategies for UXBench.}
\label{tab:ablation}
\resizebox{\columnwidth}{!}{
\begin{tabular}{l c c}
\toprule
\textbf{Training Configuration} & \textbf{ACC} & \textbf{$\Delta$ACC} \\
\midrule
Qwen3-VL-4B-Thinking (baseline) & 0.5254 & -- \\
\quad + Before Balance & Reward Hacking & -- \\
\midrule
+ Hard Negative Mining & 0.7195 & +0.1941 \\
+ Hard Neg. Mining + Positive Resampling & 0.7579 & +0.2325 \\
+ Hard Neg. Mining + Positive Augmentation & 0.7771 & +0.2517 \\
+ Hard Neg. Mining + Pos. Aug. + MultiUI & \textbf{0.7963} & \textbf{+0.2709} \\
\bottomrule
\end{tabular}
}

\end{table}

\subsection{Analysis of Asymmetric Transition Reward}

To validate the design rationale of the Asymmetric Transition Reward proposed in Section 4.2, we conduct a systematic quantitative analysis focused on the core reward mechanism. Note that this analysis uses augmented data from 8 visual understanding tasks without MultiUI integration, allowing us to isolate and evaluate the effectiveness of the transition reward mechanism independent of cross-domain training effects. We collect 2,000 reasoning samples and categorize them into ``correct'' (1,092 samples, 54.6\%) and ``incorrect'' (908 samples, 45.4\%) groups based on the correctness of final answers. For each sample, we count the number of transition markers $T$ (e.g., ``but'', ``however'').

\noindent\textbf{Significant Distributional Differences in Transition Markers.} Table~\ref{tab:transition_stats} presents key statistics on transition marker distribution. Incorrect samples exhibit an average transition marker count (14.38) nearly 6 times higher than correct samples (2.48). Most critically, 31.5\% of incorrect samples have $T>3$, compared to only 11.7\% of correct samples---a 2.7-fold difference that provides empirical support for selecting $T=3$ as our penalty threshold.

\begin{table}[t]
\scriptsize
\centering
\caption{Transition Marker Distribution Statistics for Correct vs. Incorrect Samples}
\label{tab:transition_stats}
\begin{tabular}{lccc}
\hline
\textbf{Metric} & \textbf{Correct} & \textbf{Incorrect} & \textbf{Conclusion} \\
\hline
Total Samples & 1,092 (55\%) & 908 (45\%) & - \\
Mean $T$ & 2.48 & 14.38 & \begin{tabular}[c]{@{}l@{}}Incorrect samples exhibit \\ significantly higher \\ ``overthinking''\end{tabular} \\
Median $T$ & 0 & 0 & \begin{tabular}[c]{@{}l@{}}Both distributions \\ skewed toward \\ low values\end{tabular} \\
$T=0$ Ratio & 69.0\% & 50.7\% & \begin{tabular}[c]{@{}l@{}}Correct samples tend \\ to be ``confident \\ and correct''\end{tabular} \\
$T>3$ Ratio & 11.7\% & 31.5\% & \begin{tabular}[c]{@{}l@{}}Key finding: $T>3$ is \\ a strong indicator of \\ high error rates\end{tabular} \\
\hline
\end{tabular}

\end{table}

\begin{figure}[t]
\centering
\includegraphics[width=\linewidth]{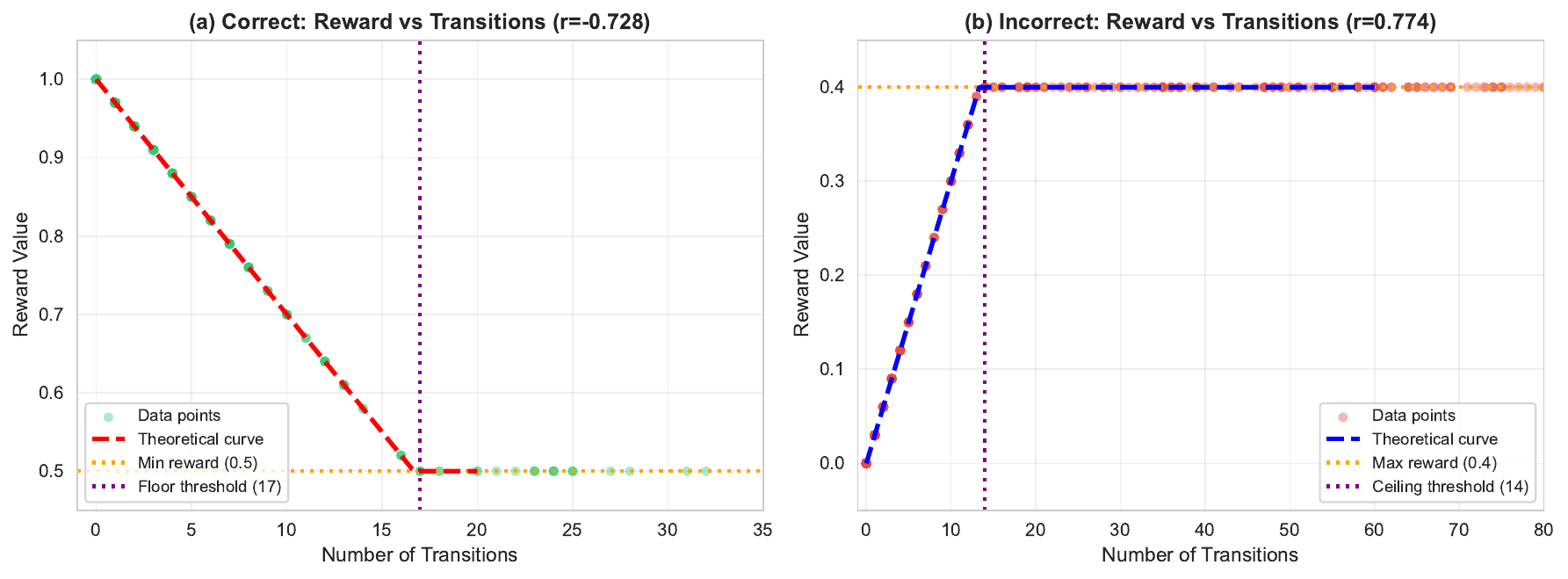}
\caption{Accuracy and Reward vs. Transition Marker Intervals. (a) Accuracy decreases monotonically as transition marker count increases; (b) Average reward exhibits a similar declining trend, validating the effectiveness of the transition marker-based reward mechanism.}
\label{fig:reward_vs_transitions}
\end{figure}

\noindent\textbf{Relationship Between Transition Markers and Accuracy.} Figure~\ref{fig:reward_vs_transitions} demonstrates that accuracy monotonically decreases from 61.0\% to 14.8\% as transition markers increase---a 4.1-fold reduction. The reward metric exhibits a parallel declining trend, validating that our transition marker-based mechanism effectively captures reasoning quality.

\begin{figure}[t]
\centering
\includegraphics[width=\linewidth]{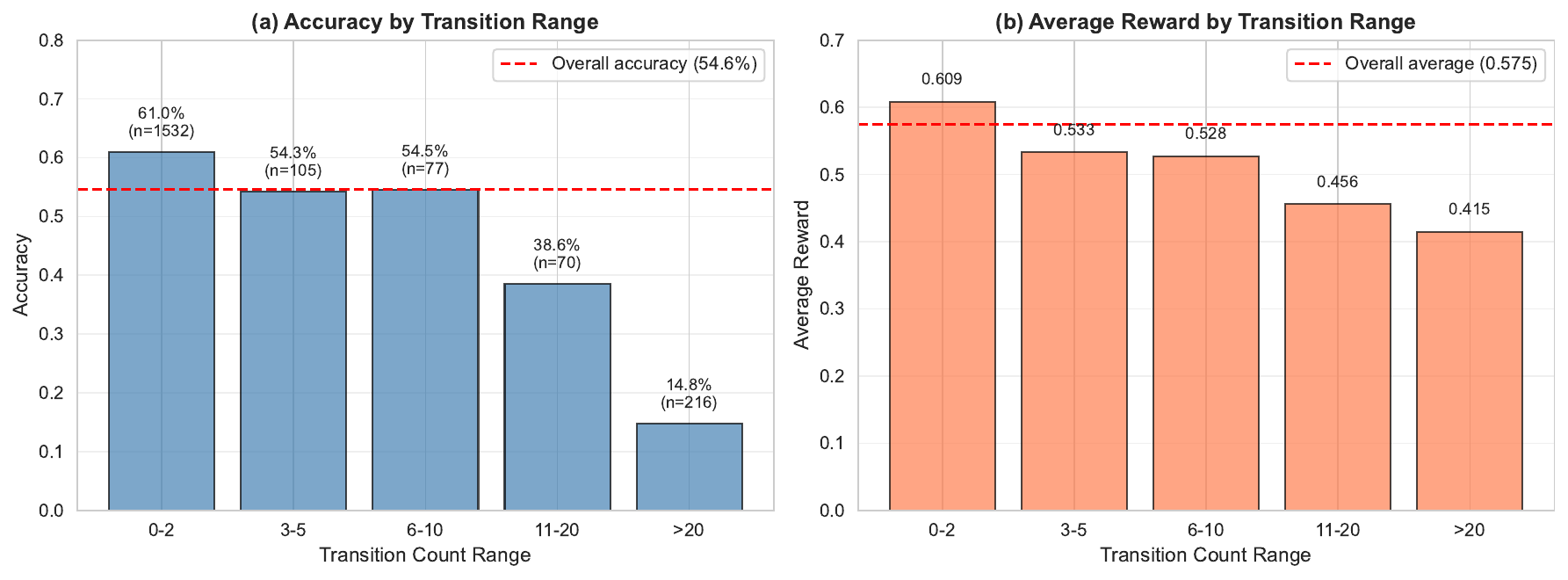}
\caption{Reward vs. Transition Markers for Correct and Incorrect Samples. (a) Correct samples: negative correlation ($r=-0.728$) between transition markers and reward; (b) Incorrect samples: positive correlation ($r=+0.774$), with upper bound at 0.4 preventing over-rewarding of verbosity.}
\label{fig:group_analysis}
\end{figure}

\noindent\textbf{Empirical Validation of the Reward Function.} Figure~\ref{fig:group_analysis} validates our asymmetric reward design. For correct samples, transition markers exhibit a strong negative correlation with reward ($r=-0.728$, $p<0.001$), matching our theoretical curve $R_{\text{correct}} = \max(1.0-0.03T, 0.5)$. For incorrect samples, the correlation is strongly positive ($r=+0.774$, $p<0.001$), following $R_{\text{incorrect}} = \min(0.03T, 0.4)$. These correlations confirm that our reward functions accurately reflect the intended design.

% \noindent\textbf{Cross-Task Consistency.} Our analysis demonstrates cross-task consistency: across all 8 tasks with accuracy ranging from 40.3\% to 72.7\%, incorrect samples consistently exhibit higher transition marker counts than correct samples, with ratios ranging from 2.1$\times$ to 16.4$\times$ (mean: 7.0$\times$). The cumulative distribution analysis reveals that at $T=3$, 88.3\% of correct samples are covered while only 68.5\% of incorrect samples are covered, representing a 19.8 percentage point gap. This clearly indicates that $T=3$ is an ideal split point, enabling targeted penalization of a significant proportion (31.5\%) of ineffective reasoning with minimal false positives (11.7\% of correct samples). This consistency validates that the reward mechanism generalizes well across diverse task types without task-specific hyperparameter tuning.

\noindent\textbf{Effectiveness Validation.} We compare our method against a baseline that provides rewards solely based on answer correctness without transition marker penalties. Our method jointly optimizes MathAccuracy and Asymmetric Transition Reward (weights 2.0 and 1.0) with a protection period for the first 40\% of training steps.

\begin{table}[t]
\footnotesize
\centering
\caption{Effectiveness validation results. 
(Acc: Accuracy, TR: Transition Reward, Len: Mean output length in tokens.)}
\label{tab:ablation}
\begin{tabular}{lccc}
\toprule
Model & Acc & TR & Len \\
\midrule
Baseline & 0.7771 & ---   & 1770 \\
Asymmetric Transition Reward & 0.7675 & 0.926 & 334 \\
\bottomrule
\end{tabular}
\end{table}

% \begin{table}[t]
% \footnotesize
% \centering
% \caption{Effectiveness Validation Results}
% \label{tab:ablation}
% \begin{tabular}{lccc}
% \hline
% \textbf{Model Variant} & \textbf{Accuracy} & \textbf{Transition Reward} & \textbf{Mean Length (tokens)} \\
% \hline

% Asymmetric Transition Reward & 80.39 & 0.926 & 334.04 \\
% Baseline & 0.7771 & - & 1770.20 \\
% \hline
% \end{tabular}
% \vspace{-1.2em}

% \end{table}

% Table~\ref{tab:ablation} validates the effectiveness of our Asymmetric Transition Reward. While maintaining competitive accuracy (76.75\% vs. 77.71\%, only -0.96\%), our method achieves an 81.1\% reduction in generation length—from 1770 to 334 tokens per sample. The high transition reward score (0.926) indicates effective reasoning control. This 5.3× reduction in output length demonstrates that our approach successfully eliminates redundant reasoning without sacrificing problem-solving capability. The baseline's excessive verbosity (1770 tokens) reflects uncontrolled overthinking, whereas our method produces concise reasoning patterns, validating that asymmetric transition penalties effectively optimize the efficiency-accuracy trade-off on UXBench.

Our method achieves comparable accuracy (77.71\% vs. 76.75\%) while reducing generation length by 81.1\% (from 1770 to 334 tokens). The transition reward of 0.926 approaches the ideal value of 1.0, demonstrating effective verbosity control. This validates that transition markers successfully identify and penalize inefficient reasoning without sacrificing accuracy.
\section{Conclusion}

We presented \textbf{UXBench}, a multimodal benchmark for fine-grained UI-based reasoning, and \textbf{UI-UX}, a reinforcement learning framework that improves MLLM performance via task-adaptive rewards and overthinking penalties. Our approach achieves strong accuracy and generalization without manual preference data.

Future work will expand UXBench with richer dimensions from cognitive psychology and real interaction logs, and integrate UI-UX into practical design assistants and automated testing, advancing AI-driven, human-centric interface evaluation.

{
    \small
    \bibliographystyle{ieeenat_fullname}
    \bibliography{main}
}


\end{document}